\pdfoutput=1

\documentclass[11pt]{article}

\usepackage[]{acl}

\usepackage{times}
\usepackage{latexsym}
\usepackage[T1]{fontenc}
\usepackage[utf8]{inputenc}
\usepackage{times}
\usepackage{latexsym}
\usepackage[inline]{enumitem}
\usepackage{amssymb}

\usepackage[T1]{fontenc}

\usepackage[utf8]{inputenc}

\usepackage{microtype}
\usepackage{graphicx}
\usepackage{multirow}
\usepackage{booktabs}

\usepackage{acronym}

\usepackage{tabularx}
\usepackage{algorithm}
\usepackage{algpseudocode}
\usepackage{subcaption}
\usepackage{array}
\usepackage{makecell}

\usepackage[skip=2pt]{caption}
\acrodef{QA}{question answering}
\acrodef{MC}{machine comprehension}

\acrodef{PeaQA}{parameter-efficient abstractive question answering}

\usepackage{xcolor}
\usepackage{array, makecell}
\usepackage[utf8]{inputenc}
\usepackage{enumitem}
\usepackage{adjustbox}
\usepackage{fancyvrb}
\usepackage{amsmath}
\usepackage{makecell}
\usepackage{svg}
\usepackage{bbm}

\title{MultiTabQA: Generating Tabular Answers\\ for Multi-Table Question Answering}

\author{
Vaishali Pal$^{1,2}$
\qquad
Andrew Yates$^1$
\qquad
Evangelos Kanoulas$^1$
\qquad
Maarten de Rijke$^1$
\\
$^1$University of Amsterdam, The Netherlands \\
\qquad
$^2$Discovery Lab, Elsevier, The Netherlands \\
\texttt{v.pal, a.c.yates, e.kanoulas, m.derijke@uva.nl}
}
\begin{document}
\maketitle

\begin{abstract}
Recent advances in tabular question answering (QA) with large language models are constrained in their coverage and only answer questions over a single table. However, real-world queries are complex in nature, often over multiple tables in a  relational database or web page. Single table questions do not involve common table operations such as set operations, Cartesian products (joins), or nested queries. Furthermore, multi-table operations often result in a tabular output, which necessitates table generation capabilities of tabular QA models. To fill this gap, we propose a new task of answering questions over multiple tables. Our model, MultiTabQA, not only answers questions over multiple tables, but also generalizes to generate tabular answers. To enable effective training, we build a pre-training dataset comprising of 132,645 SQL queries and tabular answers. Further, we evaluate the generated tables by introducing table-specific metrics of varying strictness assessing various levels of granularity of the table structure. MultiTabQA outperforms state-of-the-art single table QA models adapted to a multi-table QA setting by finetuning on three datasets: Spider, Atis and GeoQuery.
\end{abstract}

\acresetall

\section{Introduction}
Question answering (QA) over multiple tables aims to provide exact answers to natural language questions with evidence from one or more tables \cite{https://doi.org/10.48550/arxiv.2207.05270}.  This is in contrast to single-table QA, which has been the focus of tabular QA research to date~\citep{liu2021tapex,nan2021feta,zhu2021tatqa,herzig-etal-2020-tapas}. Even though groups of related tables are ubiquitous in real-world corpora, such as relational databases or tables in a web page, multi-table QA remains a largely unexplored area. 
\begin{figure}[th!]
 \centering
    \resizebox{0.97\columnwidth}{!}{\includegraphics[width=1\columnwidth]{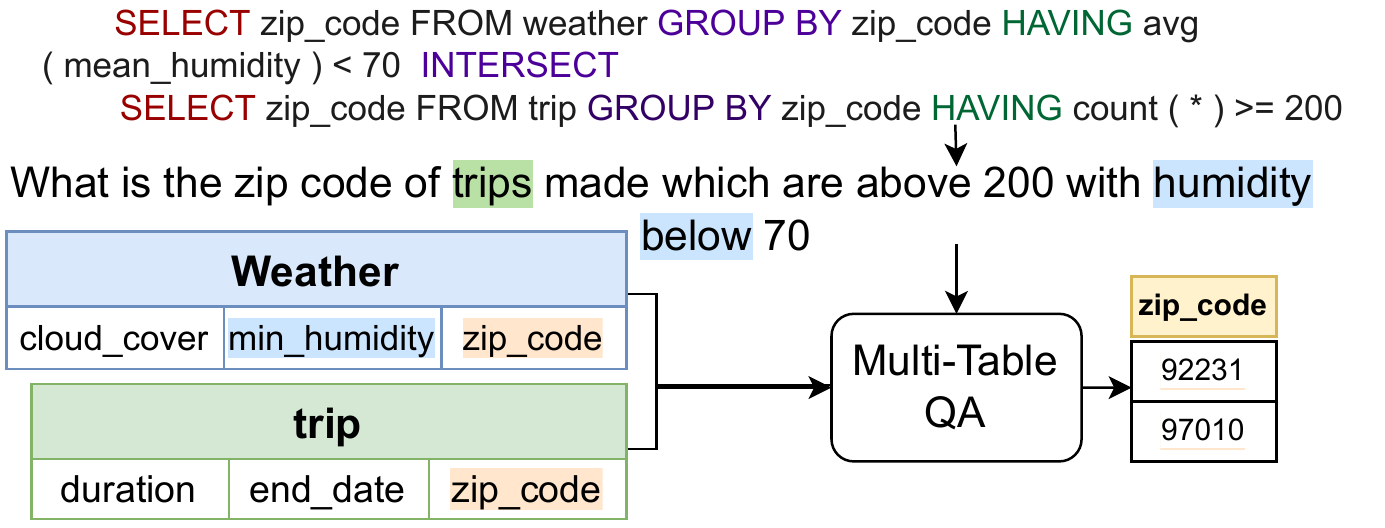}}
 \caption{Multi-table QA. The QA model generates a tabular answer from either a natural language question or an SQL query and one or more tables as input context.}
 \label{fig:main_diagram}
\end{figure}
To address this gap, we propose a new task of answering questions over multiple tables. Our multi-table QA model, MultiTabQA,\footnote{Code and data are at: \url{https://github.com/kolk/MultiTabQA}}
addresses novel challenges introduced by multi-table context. These include complex queries involving chains of reasoning, disambiguation of relevant table names at each reasoning step, and generating a final table as answer. 
It also leads to novel question-types that are unnatural in a single-table setting. 
For instance, questions involving operations specific to multiple tables, such as Cartesian products (\textit{outer joins}, \textit{inner joins}) and set operations (such as \textit{intersect}, \textit{union}, \textit{in}), are unique to and common in a multi-table scenario. 
Furthermore, such multi-table operations often result in a tabular answer and they necessitate table generation capabilities of the QA model. 

Figure~\ref{fig:main_diagram} depicts an example of a question involving two tables, \textit{I would like to know the zip code of trips taken above 200 with humidity below 70}, and its associated input tables, \textit{Weather} and \textit{trip}. 
A multi-table QA model is expected to disambiguate records from different tables (the question phrase \textit{zip code of trips} grounds the column \textit{zip\_code} of Table~\textit{trip}; the question phrase \textit{humidity below 70} grounds column \textit{min\_humidity} of Table \textit{Weather}), learn associations among inter-table columns (\textit{zip\_code} in both tables) and intra-table columns (\textit{min\_humidity} and \textit{zip\_code} in the \textit{Weather} table), and finally compute the required operations (\textit{intersect}, \textit{count}) and generate the tabular answer.

Recent work on tabular QA can be categorized into two major directions:
\begin{enumerate*}[label=(\roman*)]
\item Semantic parsing-based techniques \cite{DBLP:journals/corr/PasupatL15,zhongSeq2SQL2017,Cai2022STARSG}, which have been the dominant approach to answering multi-table complex questions. Such methods transform a natural question to a logical form, which is used to query a relational database to extract the answer. However, these techniques are restricted to relational databases and cannot be applied to tables from other sources such over web tables, tables in text documents, and any non-normalized tables. Additionally, they require expensive and expert human annotations \cite{yu-etal-2018-spider,KaggleDBQA2021} formulating SQL queries from natural questions. 
\item Modeling the problem as a sequence generation/classification task \cite{tabert2020,Zhang:2020:SET,herzig-etal-2020-tapas, zhu2021tatqa, liu2021tapex,Cheng2021HiTabAH,nan2021feta,Ma2021OpenDQ,pal-etal-2022-parameter,https://doi.org/10.48550/arxiv.2207.05270}, where an end-to-end trained neural model is not only responsible for question/query understanding but also table reasoning. 
Existing end-to-end neural models are either classification-based \cite{herzig-etal-2020-tapas,zhu2021tatqa}, where the model detects the answer span and classifies one table operator associated with the span, or they are sequence generation-based \cite{nan2021feta,Zhang:2020:SET,liu2021tapex}, where the model generates the answer as a span of text in an auto-regressive manner.
\end{enumerate*}

Our work focuses on the latter direction of research. We train a neural model to mimic a semantic parser and generate the answer. A clear distinction of our work compared to existing end-to-end models is that our proposed model, MultiTabQA, does not operate in the constrained setting of a single input table, but can accommodate one or more tables in the input and the associated multi-table operators. Additionally, MultiTabQA performs the task of structured table generation, which imposes structure aspects to the generated output such as table schemas, alignments of rows and columns, relationships between column-headers and column values. Generating structured tables as output requires table-specific evaluation metrics which we define and use to evaluate the generated tables.
To effectively train the model, we generate a pre-training dataset with multi-table SQL queries and tabular answers built over an existing semantic parsing dataset, Spider \cite{yu-etal-2018-spider}. Our dataset consists of $132,645$ samples comprising of SQL queries, associated natural language questions, input tables, and tabular answers. 
To the best of our knowledge, this is the first work to address the task of multi-table QA and generate tabular output.

Our main contributions can be summarized as:
\begin{enumerate}[label=(\arabic*),leftmargin=*,nosep]
    \item We fill-in the gap of existing tabular QA methods, which operate only on single tables, by proposing a new task of answering questions over multiple tables. Our work increases the breadth of question types that can be handled by single tabular QA methods.
    \item Our proposed multi-table QA model generates structured tables imposed by multi-table operations. Table generation introduces generation challenges such as maintaining row-column alignment, table-header generation, etc. 
    \item We release a multi-table pre-training dataset comprising of $132,645$ samples of SQL queries and tabular answers.
    \item We introduce table generation metrics that capture different levels of granularity and strictness to evaluate our proposed model.
\end{enumerate}

\begin{figure*}[h!]
 \centering
    \resizebox{0.97\textwidth}{!}{\includegraphics[width=1\textwidth]{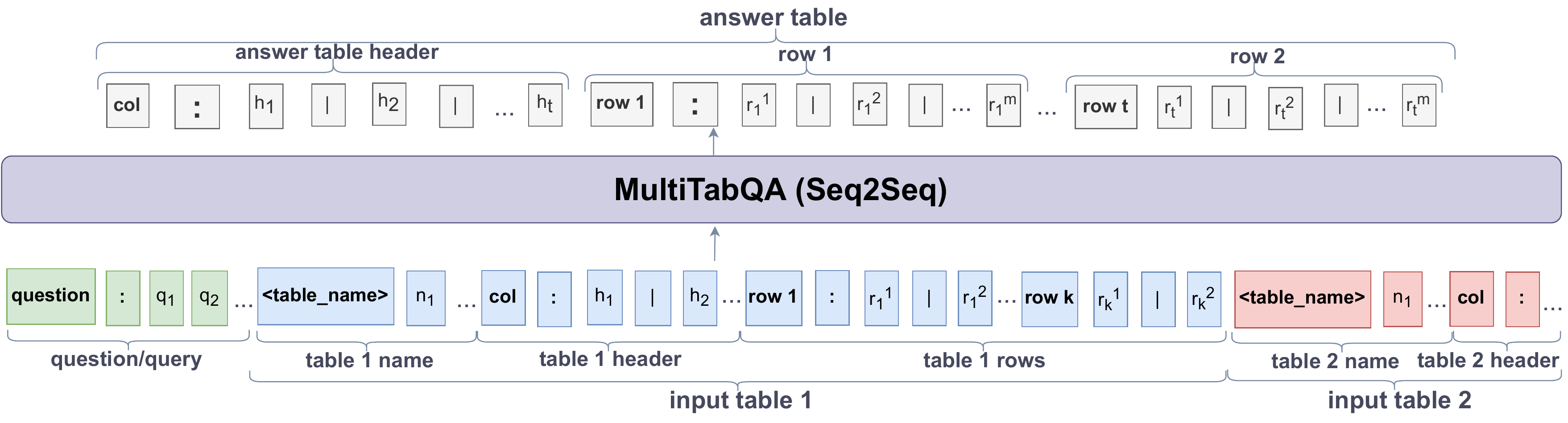}}
    \caption{Architecture of MultiTabQA model. Given a natural language question/SQL query and the associated tables as an input sequence, the seq2seq model performs tabular reasoning and generates a tabular answer. Start of an input table is identified with keyword \textbf{<table\_name>} which also indicates that the next tokens comprises the table name. \textbf{col:} indicates that the next tokens are table headers. Rows in a table are identified with keyword \textbf{row i:}, columns are separated by |.}
 \label{fig:architecture}
\end{figure*}

\section{Methodology}
We frame multi-table question answering as a sequence-to-sequence task and train an auto-regressive transformer encoder-decoder model to generate the tabular answer. Given a question $Q$ consisting of a sequence of $k$ tokens $q_1, q_2, \ldots, q_k$ and a set of $N$ tables, $T_N = \{t_1, t_2, \ldots, t_n\}$, the goal of the multi-table QA model is to perform chains of \emph{operations} over $T_N$, constrained by $Q$, and generate a table $T_{out}$. The model always generates a table, $T_{out}$, which can be single celled for scalar answers, single rowed or columned for list-based answers, and multiple rows and columns for tabular answers. In all cases, the model also generates column headers revealing important semantics associated with the generated values.  

\paragraph{Training approach.} We follow a Curriculum Learning approach \cite{10.1145/1553374.1553380} by sequentially increasing the complexity of tasks to train MultiTabQA. The first stage of training is a pre-training step where the training objective is two-fold: 
\begin{enumerate*}[label=(\roman*)]
\item learn to generate correct tabular answers from SQL, and
\item understand the associations between related input tables. 
\end{enumerate*}
The final training stage is fine-tuning where the model learns to understand natural language questions with their inherent ambiguity in addition to retaining its ability of reasoning over tables and generating a tabular answer. We discuss the training process in detail in Section~\ref{sec:training}.

\paragraph{Model input/output.} The input to the model is a sequence comprised of the query or the natural language question, followed by a sequence of input tables, represented by the table name and the corresponding flattened table. Table names are important for disambiguating tables in multi-table QA setting. Specifically, the input sequence is represented as
$question\ [table_1\ rep]\ [table_2\ rep] \ldots [table_n\ rep]$ where $[table_i\ rep]$ is the representation of the $i$-th table. As depicted in Figure \ref{fig:architecture}, the $i$-th table is flattened in row-major format and represented as 
\begin{multline*}
\textbf{<table\_name>:}\ n_1\  n_2\ \mid \textbf{col:}\ h_1 \mid h_2 \mid \ldots \mid h_k\\
\textbf{row\ 1:}\ r^1_{1} \mid \ldots \mid r_1^{m} \ldots \textbf{row\ k:}\ r^1_{k} \mid \ldots \mid r_k^{m},
\end{multline*}
where $n_1,\ldots, n_2$ is the sequence of table name tokens, $h_j$ is $j$-th column header, $r^i_m$ is the $i$-th row and $m$-th column cell. The boldface words are keywords specifying semantics of the next tokens. The output of the model is also a flattened table in row-major format, i.e., $[table_{ans}\ rep]$, but without a table name. As depicted in Figure~\ref{fig:architecture}, the generated table, $[table_{ans}\ rep]$, is of the form:
\begin{multline*}
\textbf{col:}\ h_1 \mid h_2 \mid \ldots \mid h_k\ \textbf{row\ 1:}\ r^1_{1} \mid \ldots \mid r_1^{m} \\
\textbf{row\ 2:}\ r^1_{2} \mid \ldots\mid r_2^{m} \ldots \textbf{row\ k:}\ r^1_{k} \mid \ldots\mid  r_k^{m}.
\end{multline*}


\section{Dataset}
\label{section:dataset}
To effectively train a multi-table QA model, the dataset needs to cover three aspects:
\begin{enumerate*}[label=(\roman*)]
\item multi-table context, 
\item tabular answers, and
\item natural questions. 
\end{enumerate*}
Given the absence of large-scale datasets covering all three aspects,  we transform existing semantic parsing and single-table QA datasets to focus on a single aspect before training with samples covering all three aspects. 

\subsection{Single table pre-training dataset}
\label{sec:tapex_pretraining}
One of the sub-tasks of pre-training is to generate tabular answers. We hypothesize that tuning the model to generate tables may lead to a warm-start and better convergence in a multi-table QA setting. To enable such experiments, we modify the large-scale single-table QA Tapex pre-training dataset~\citep{liu2021tapex} to accommodate tabular answers. The dataset contains $1,834,419$ samples of query, input table and factoid answers. The tables in the dataset are not named as there is no need for table disambiguation in a single table setting. The SQL queries are semi-formal (do not contain the \texttt{FROM} clause with a table name) and cannot be used to query a real SQL database. We insert a placeholder table name in the queries and the corresponding input tables to extract the tabular answer from the database. Transforming the factoid answers to tables leads to single-celled or single-rowed tables. The modified dataset helps the model to understand simple tables and reason over semi-formal queries to generate simple tables.

\subsection{Multi-table pre-training dataset}
\label{sec:multitab_pretraining_data}
We develop a multi-table pre-training dataset over the database of Spider \cite{yu-etal-2018-spider}. Spider is a cross-domain complex semantic parsing dataset for text-to-SQL translation. It consists of $10,181$ questions and $5,693$ SQL queries. The questions are over $200$ databases of multiple tables covering 138 different domains. The training, development and test splits do not contain overlapping databases to test a model's generalizability to new databases.

We first adapt the existing samples of Spider for our task. We use the ground-truth SQL queries of Spider as input query for pre-training over multiple tables. We automatically extract all input table names from the SQL query and retrieve the input tables\footnote{We use SQLite3 and pandas for extracting tables.} from the relational database. The query, extracted table names, and retrieved tables are inputs to our multi-table QA model. We extract the answer table with the SQL query by querying the relational database. Answer table headers reveal important semantics of the associated column values such as the numeric operation (\textit{average}, \textit{sum}, etc.), numeric scales (million, thousand, kms, meters, etc.), or entity facets (name, date, etc.).  This process generates $3816$ samples comprising of \emph{query}, \emph{question}, \emph{table\_names}, \emph{tables} and \emph{answer}. 

We further augment the modified Spider dataset with $132,645$ samples of synthetic queries. This leads to an augmented multi-table pre-training dataset of $136,461$ unique training samples comprising of $3816$ Spider samples and $132,645$ synthetic samples. The validation set comprises of $536$ samples from the Spider validation set pre-processed as described above to adapt to our task. 

Existing work on semantic parsing \cite{Shi:Zhao:Boyd-Graber:Daume-III:Lee-2020,yu2021grappa} have utilized hand-crafted templates to generate large-scale corpora of synthetic queries, but are constrained in their coverage with no multi-table operations \cite{Shi:Zhao:Boyd-Graber:Daume-III:Lee-2020} or limited coverage with no table \emph{joins} and lacking diversity in \emph{set} operations \cite{yu2021grappa}. This motivates us to generate our augmented pre-training dataset for multi-table QA using multi-table SQL templates. 

Our synthetic queries are generated from $45$ manually crafted templates over the Spider database and hand-crafted rules for operation types. The query templates have placeholders for aggregation, relational operations, table name and headers which are randomly assigned during query generation process. For example, to generate multi-table \textit{join} queries, we instantiate the templates by randomly choosing tables from a database with at least one common header. For \textit{set} operations, all tables participating in a multi-table query requires all table headers to match. We design SQL templates in increasing order of complexity starting with simple SQL templates and progressively adding components which increases its complexity. For example, for single-table queries, we use the simplest template \emph{"SELECT * FROM \{table\_name\}"} whereas for multi-table templates such as \emph{joins}, the simplest template is \emph{"SELECT T1.\{table1\_cols\}, T2.\{table2\_cols\} FROM \{table\_name1\} as T1 JOIN \{table\_name2\} as T2 ON T1.\{common\_col\} = T2.\{common\_col\}"}. We progressively add SQL components such as aggregations, \emph{where} conditions, \emph{group by} and \emph{having} clauses to generate templates of increasing complexity. This process results in $14$ templates for \emph{joins}, $4$ templates for each set operation: \emph{intersect}, \emph{union} and \emph{except}. To avoid catastrophic forgetting for single table queries, we also instantiate $14$ single-table queries with increasing complexity.


\paragraph{Quality control.} We ensure correctness of the synthetic samples by discarding SQL queries that executes to an error or empty table. We also apply the process on the modified Spider, Atis and GeoQuery data to discard SQL query and the corresponding natural language question to ensure that all questions are answerable. 


\subsection{Multi-table QA dataset}
\label{sec:multitabQAdata}
We fine-tune and evaluate our model on the natural language questions of semantic parsing datasets: Spider, GeoQuery~\cite{10.5555/1864519.1864543}, and Atis~\cite{price-1990-evaluation,dahl-etal-1994-expanding}. GeoQuery is a semantic parsing dataset to query into a database of United States geography.\footnote{This data is made available under under GPL 2.0 license.} Atis is a semantic parsing dataset\footnote{This data is made available under MIT license.} with a collection of $4,379$ questions, corresponding SQL queries and a relational database to a flight booking system \cite{nl2sql2017}. Similar to the Spider dataset processing described in Section \ref{sec:multitab_pretraining_data}, we first extract the input table names from the available SQL queries and query the relational database for the input tables.\footnote{We preprocess the Atis and GeoQuery data samples available at \url{https://github.com/sriniiyer/nl2sql/tree/master/data}.} We also extract the tabular answers using the SQL queries. We discard any samples that executes to an error or empty table. We use the corresponding natural language question for each SQL query as the user utterance for fine-tuning.  This results in $6,715$ training samples and $985$ validation samples for Spider.  We also process the $600$ GeoQuery samples provided in \cite{nl2sql2017} to create a subset of $530$ training samples, $49$ validation samples and $253$ test samples. We process and generate an Atis subset of $384$ training samples, $45$ evaluation samples and $86$ test samples. We discard Atis queries with very large input tables (with > $10,000$ rows). This restriction enables us to correctly evaluate question answering capabilities of a model by ignoring samples with truncated input sequences including entire input tables from the second table onward. Truncation of tables leads to incorrect answers for any numeric operation such as \textit{average}, \textit{intersect} and the evaluation scores would no longer reflect reasoning capabilities of the model. 

\section{Training}
\label{sec:training}
\begin{figure}[t!]
 \centering
    \resizebox{0.99\columnwidth}{!}{\includegraphics[width=1\columnwidth]{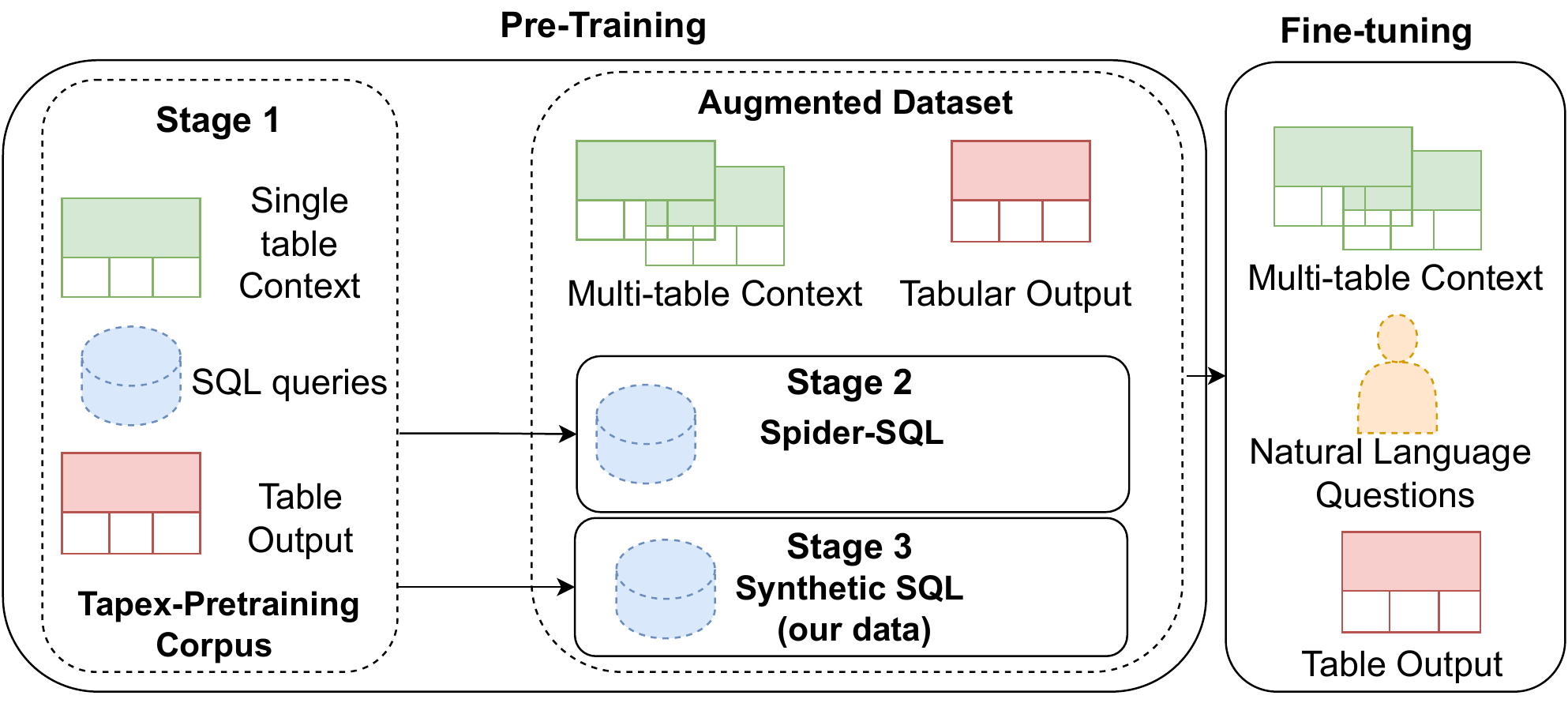}}
    \caption{Four stage training procedure. The first three stages are pre-training, followed by fine-tuning.}
 \label{fig:training}
\end{figure}

We follow a curriculum learning approach by sequentially training the model on sub-tasks of increasing complexity as depicted in Figure~\ref{fig:training}.
Broadly, we first pre-train the seq2seq model to mimic an SQL parser and further fine-tune it on the downstream multi-table QA task. Pre-training the model on unambiguous SQL queries leads to better convergence and warm-start for the closely related downstream multi-table QA task. We further segregate the pre-training by first addressing the simpler sub-task of generating tables from single table queries. This is immediately followed by pre-training on multi-table query answering where complex SQL queries are utilized to train the model to learn multi-table associations from unambiguous complex queries, reason over the tables and generate tabular answer. The final stage of training is the downstream multi-table QA from natural language questions. Natural language introduces ambiguity, ellipses and co-references which increases complexity and is thus the final stage of training. For each stage,  we choose the model with the best table exact match accuracy on the corresponding validation set, defined in Section \ref{sec:eval_metric}, as the initialization for training the next stage.

\subsection{Pre-training}
\label{sec:pretraining}
Pre-training of MultiTabQA is conducted in two stages in a curriculum learning fashion: Stage~1 is single table QA where the model learns to generate tabular answers from relatively simple SQL queries. Stage~2 is multi-table QA where the model trained in Stage~1 is further tuned for multi-table SQL QA. 

\paragraph{Stage 1.} We first train MultiTabQA on the task of generating tables from SQL queries over single tables. The tabular answer to be generated is simple and single-columned. For this stage, we use the modified Tapex pre-training corpus described in Section~\ref{sec:tapex_pretraining}. We train the model on $1,834,419$ samples for two epochs. This stage provides a good initialization for multi-table QA in next stages. 

\paragraph{Stage 2 + Stage 3.} We further pre-train the model on multi-table QA. For this, we tune our model on SQL queries from the modified Spider and synthetic dataset. We call tuning with only the modified Spider SQL samples \textit{Stage~2}, and tuning with only the synthetic dataset \textit{Stage 3}. We utilize the larger augmented dataset comprising of the modified Spider SQL (Stage~2) and our synthetic samples (Stage~3) as described in Section \ref{sec:multitab_pretraining_data} to train the final pre-trained model for $30$ epochs. We call this setting \textit{Stage~2+3}. We compare these three multi-table pre-training settings in Section~\ref{sec:results}.

\subsection{Fine-tuning}
The final stage of training is fine-tuning the pre-trained model on natural language questions. Natural questions are ambiguous compared to formal SQL and used at the last stage of training. We fine-tune the pre-trained model on the $6,715$ natural questions, extracted input and output tables for Spider as described in Section \ref{section:dataset} and evaluate on $985$ samples of the validation set. To observe the performance of the pre-trained model on out-of-domain database tables, we also fine-tune the pre-trained model on Atis and GeoQuery datasets. For all the fine-tuning datasets, we train for $60$ epochs.
\begin{table*}[t!]
\centering
\resizebox{0.97\textwidth}{!}{
\begin{tabular}{ll c ccc ccc ccc}
\toprule
\multirow{2}{*}{\textbf{Dataset}} & \multirow{2}{*}{\textbf{Model}} & \multirow{2}{*}{\textbf{\makecell{Table\\EM (\%)}}} &  \multicolumn{3}{c}{\textbf{Row EM (\%)}} &  \multicolumn{3}{c}{\textbf{Column EM (\%)}} & \multicolumn{3}{c}{\textbf{Cell EM (\%)}} \\
\cmidrule(r){4-6}
\cmidrule(r){7-9}
\cmidrule{10-12}
& & &  \textbf{P} & \textbf{R} & \textbf{F1} & \textbf{P} & \textbf{R} &  \textbf{F1} & \textbf{P} & \textbf{R} & \textbf{F1}\\
\hline
\multirow{2}{*}{Spider} & \texttt{tapex-base} & 18.99 & 17.28 &	 19.83 & 18.27 & 19.75 & 19.39 & 19.57 & 23.15 & 27.71 & 25.03 \\
& MultiTabQA &  \textbf{25.19}\rlap{*} & \textbf{22.88}\rlap{\dag} & \textbf{24.64}\rlap{*} &	\textbf{23.70}\rlap{*} & \textbf{26.86}\rlap{*} & \textbf{26.76}\rlap{*} & \textbf{26.81}\rlap{*} & \textbf{28.07}\rlap{\dag} &	\textbf{31.23}\rlap{*} & \textbf{29.55}\rlap{*}  \\
\midrule
\multirow{2}{*}{GeoQ} & \texttt{tapex-base} & 39.84 & 22.43 & 30.74 & 24.89 & 39.48 & 39.76 & 39.62 & 21.98 & 30.88 & 24.67 \\
& MultiTabQA &  \textbf{52.22}\rlap{*} &  \textbf{72.39}\rlap{*} &  \textbf{46.90}\rlap{*} &  \textbf{41.38}\rlap{*} &  \textbf{52.10}\rlap{*} &  \textbf{52.22}\rlap{*} &	 \textbf{52.16}\rlap{*} &  \textbf{37.16}\rlap{\dag} &  \textbf{46.92*} &  \textbf{41.33}\rlap{*}  \\
\midrule
\multirow{2}{*}{Atis} & \texttt{tapex-base} & 72.20 & \textbf{57.07}\rlap{\dag} & 57.69 & \textbf{55.08} & \textbf{72.20}\rlap{\dag} & 72.20 & 72.20 & \textbf{57.07}\rlap{\dag} & 57.69 & \textbf{54.48} \\
& MultiTabQA& \textbf{73.88}\rlap{\dag} & 38.29 & \textbf{92.19}\rlap{*} & 54.36 & 69.55 & \textbf{75.24}\rlap{\dag}	& \textbf{72.29} & 38.16 & \textbf{92.56}\rlap{*} & 54.16 \\
\bottomrule
\end{tabular}
}
\caption{Average scores of models fine-tuned on 5 different seeds with Multitable-Natural Questions (NQ) datasets. \texttt{tapex-base} is used as baseline while \texttt{MultiTabQA} is our fine-tuned model. Table EM indicates table exact match accuracy. For all other table units (row, column, and cell), P is Precision, R is Recall, and F1 is F1 score for exact match metric. An (*) denotes significance at p < 0.005 and an (\dag) denotes a significance at p < 0.05 for t-test.}
\label{tab:exact_match_finetuning_results}

\end{table*}

\section{Evaluation metrics}
\label{sec:eval_metric}
While denotation accuracy has been widely used in semantic parsing \cite{DBLP:journals/corr/PasupatL15,zhongSeq2SQL2017,Cai2022STARSG}, it is not directly applicable for our task where tabular input encoding, reasoning, and generation are performed by the same model. Evaluating the answer table not only requires matching the generated values but also the table structure. Moreover, tables store factual information such as named entities, dates, numbers, etc in an ordered manner. This makes lexical metrics measuring surface-form overlap more suitable than semantic metrics measuring the underlying meaning of paraphrased sequences. Moreover, table components such as rows, columns and cells are standalone units which capture different levels of semantics and relationships with the surrounding table component. For example, rows capture data records while columns capture the features of each record. Cells capture the lowest level of self-contained facts and requires complete match with the target. For example, a cell with the entity ``United Kingdom'' should not be partially matched with the predictions ``United Nation'', ``United'' or ``Kingdom''. Similarly, a numeric value such as ``123.45'' should not be partially matched with ``12.45'', ``23.45'' or ``12''. Numeracy pose a challenge to seq2seq models \cite{nogueira2021investigating,Pal2021InvestigatingNL} specially in the extrapolation setting where semantic match of unseen numbers may not be an ideal. Considering all these factors, we focus on lexical match to measure model effectiveness. 
 
\paragraph{Table exact match.} We define \emph{table exact match Accuracy Table EM)} as the percentage of predicted tables which exactly matches the target tables. Table exact match evaluates ordering of rows, columns and table headers and exact lexical matching of table values. It is a strict binary measure which treats partial matches as incorrect. However, many queries do not impose ordering among columns or rows, and strict table exact match may not be the ideal indication of model efficacy. To measure partial correctness, we treat rows, columns and cells as units at varying levels of granularity which have ordered associations among the values within the unit. We evaluate partial correctness with exact match of rows, columns and cells.

\paragraph{Row exact match.} To relax the strict criterion of table exact match, we first measure correctness on table rows. Row exact match do not consider ordering of rows in the generated table but requires ordering of values within the row. We define a correctly generated row to be a predicted row that exactly matches any target rows in the target table. \emph{Row exact match precision} is the percentage of correctly generated rows among all the predicted rows in the evaluation dataset. \emph{Row exact match recall} is the percentage of correctly generated rows among all the target rows in the evaluation dataset. 

\paragraph{Column exact match.}  Unlike rows, which represent records in relational databases, columns represent attributes where column header provides semantic meaning to the values. Hence, a correct column is defined as a generated column that first exactly matches a target column header and further the column values. Column exact match measures ordering of values within a column. \emph{Column exact match precision} is the percentage of correctly generated columns among all generated columns in the evaluation set.  \emph{Column exact match recall} is the percentage of correctly generated columns among all target columns in the evaluation set.  

\paragraph{Cell exact match.} \emph{Cell exact match} is the most relaxed measure of model efficacy at the lowest level of granularity (cells) where table structure is not measured. A cell is correct if it matches any cell in the corresponding target table. \emph{Cell exact match precision} is the percentage of correctly predicted cells among all predicted cells in the dataset. \emph{Cell exact match recall} is the percentage of correctly predicted cells among all target cells in the dataset.

\begin{figure}[t!]
 \centering
    \resizebox{0.97\columnwidth}{!}{\includegraphics[width=1\columnwidth]{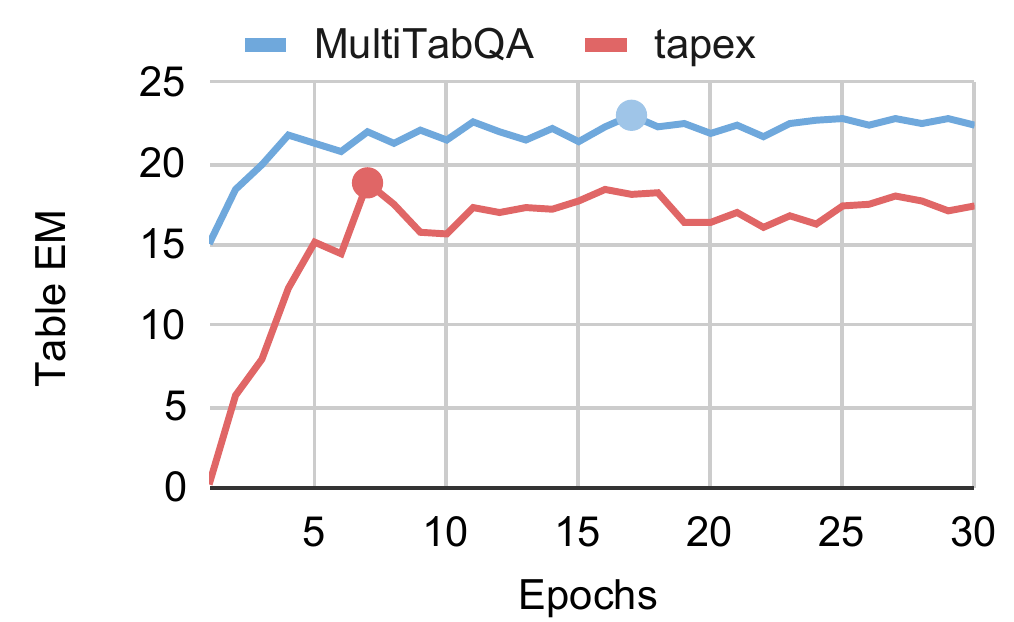}}
    \caption{Validation table exact match scores of \texttt{MultiTabQA} vs.\ \texttt{tapex-base} on Spider evaluation set natural language questions during fine-tuning. The points are highest validation scores for each model.}
 \label{fig:training_progress}

\end{figure}

\section{Experimental setup and results}
\label{sec:results}
We use \texttt{tapex-base} \cite{liu2021tapex} as the base model for all our experiments.  \texttt{tapex-base} is a single table question answering model (140M parameters) trained to approximate table reasoning by pre-training to mimic an SQL parser. For both the pre-training and fine-tuning process, we use a batch size of $8$ and gradient accumulation of $32$ to emulate an effective batch size of $256$, a learning rate is $1e^{-9}$. The maximum sequence length of both encoder and decoder is set to $1024$. We run all our pre-training experiments on four A6000 48GB GPUs and fine-tuning on one A6000 GPU. 

We observe from Figure~\ref{fig:training_progress} that the three stage pre-training leads to a warm-start for fine-tuning and better convergence compared to the baseline \texttt{tapex-base}. For our experiments, we compare the effectiveness of the MultiTabQA model with fine-tuned \texttt{tapex-base} on the $6,715$ natural questions from Spider. The fine-tuned \texttt{tapex-base} acts as baseline for studying the adaptability of state-of-the-art single table model to a multi-table setting.  We report the mean scores of 5 training runs initialized with different seeds in Table~\ref{tab:exact_match_finetuning_results}. We conduct statistical significance test (t-test) on the mean scores of the 5 runs and report the significance with $p < 0.05$ and $p < 0.005$.  We observe that our multi-stage training process leads to improvement in scores on all table exact match accuracy across all datasets compared to fine-tuned \texttt{tapex-base}. The difference in table exact match is highest for GeoQuery where MultiTabQA outperforms \texttt{tapex-base} by $12.38\%$, Spider by $6.20\%$ and Atis by $1.68\%$. For F1 and Recall scores on row, column and cell exact match, a similar pattern is observed where MultiTabQA outperforms \texttt{tapex-base} on all datasets. MultiTabQA outperforms \texttt{tapex-base} by $5.43\%$ on row F1, $7.24\%$ on column F1, and $4.52\%$ on cell F1 for Spider. On GeoQuery, MultiTabQA outperforms by $16.49\%$ on row F1, $12.54\%$ on column F1 and $16.66\%$ on cell F1 scores. All results on Spider and GeoQuery are significant with a p-value less than a critical value of $0.05$ indicating strong evidence that MultiTabQA is a superior model. On Atis, we observe that MultiTabQA underperforms on precision but outperforms on recall by a large margin. The difference in recall is larger than precision indicating that MultiTabQA generates more target rows, columns and cells of Atis correctly (higher recall) and hallucinates spurious rows and cells (lower precision). However, the F1 scores are better for MultiTabQA. \texttt{tapex-base} is unable to correctly generate target rows, cells and columns (lower recall), but the few generated ones are correct (higher precision). The low number of test samples (85) of Atis and variations in the hallucinations in different runs makes the precision scores statistically non-significant. However, the recall scores provide very strong evidence ($p < 0.005$) of the superiority of MultiTabQA in generating correct table units compared to \texttt{tapex-base}.

\begin{table*}[t!]
\centering
\resizebox{0.97\textwidth}{!}{
\begin{tabular}{ll c ccc ccc ccc}
\toprule
\multirow{2}{*}{\textbf{\makecell{Pre-training \\ stages}}} & \multirow{2}{*}{\textbf{\makecell{Query\\type}}} &  \multirow{2}{*}{\textbf{\makecell{Table\\EM(\%)}}} &  \multicolumn{3}{c}{\textbf{Row (\%)}} &  \multicolumn{3}{c}{\textbf{Column (\%)}} & \multicolumn{3}{c}{\textbf{Cell (\%)}} \\
\cmidrule(r){4-6}
\cmidrule(r){7-9}
\cmidrule{10-12}
& & &  \textbf{P} & \textbf{R} & \textbf{F1} & \textbf{P} & \textbf{R} &  \textbf{F1} & \textbf{P} & \textbf{R} & \textbf{F1}\\
\midrule
2  & \multirow{3}{*}{SQL} & $21.46$ & $18.60$ & $18.88$ & $18.74$ & $21.98$ & $21.90$ & $21.94$ & $24.19$ & $25.89$ & $25.01$\\
1+2  & & $20.52$ & $14.13$ & $20.06$ & $16.58$ & $18.87$ & $20.87$ & $19.82$ & $19.24$ & $25.83$ & $22.05$ \\
1+2+3 & & $\textbf{29.10}$ & $\textbf{23.15}$ & $\textbf{25.62}$ & $\textbf{24.32}$ & $\textbf{31.66}$ & $\textbf{31.50}$ & $\textbf{31.58}$ & $\textbf{29.95}$  & $\textbf{32.92}$  & $\textbf{31.36}$ \\
\midrule
2 & \multirow{3}{*}{NL} & $19.41$ & $16.51$ & $19.48$ & $17.87$ & $20.13$ & $20.11$ & $20.12$ & $21.12$ & $26.55$ & $23.52$ \\
1+2 & &  $20.12$ & $11.67$ & $21.09$ & $15.03$ & $19.54$ & $19.97$ & $19.76$ & $16.26$ & $29.22$ & $20.90$ \\
1+2+3 & &  $\textbf{24.49}$ & $\textbf{24.95}$ & $\textbf{24.87}$ & $\textbf{24.91}$ & $\textbf{26.80}$ & $\textbf{26.91}$ & $\textbf{26.86}$ & $\textbf{28.44}$ & $\textbf{31.06}$ & $\textbf{29.69}$ \\	
\bottomrule
\end{tabular}
}
\caption{Ablation on datasets in our multi-stage pre-training processes for 1 run of experiments. The two sections show scores for different question types: SQL queries (top) and natural language (NL) questions (bottom). In a section each row shows a training process with different stages: Pre-training on Stage 2, pre-training on Stages 1+2, and all pre-training Stages 1+2+3. Table EM is table exact match accuracy; P is Precision; R is Recall; and F1 is F1 score for exact match of row, column, and cell.}
\label{tab:exact_match_pretraining_results}
\end{table*}

\paragraph{Qualitative analysis.} Multi-table QA models must perform numeric reasoning, understand multi-table schemas and comprehend natural language. A success case also depicts this. For the question \emph{how many likes does kyle have?} with 2 input tables:
\begin{center}
\begin{minipage}[c]{0.48\columnwidth}
\centering
\small
\captionsetup{labelformat=empty}
 \captionof{table}{highschooler}
\begin{tabular}{|c|c|c|}
\hline
id &  name & grade  \\
\hline
1510 & jordan & 9\\
\ldots & \ldots & \ldots  \\
\textbf{1934} & \textbf{kyle} & 12  \\
1661 & logan & 12 \\
\hline 
\end{tabular}
\end{minipage}
\begin{minipage}[c]{0.48\columnwidth}
\small
\captionsetup{labelformat=empty}
 \captionof{table}{likes}
\centering
\begin{tabular}{|c|c|}
\hline
student\_id & like\_id  \\
\hline
 1689 & 1709  \\
 \ldots & \ldots  \\
 1501 & \textbf{1934} \\
 \textbf{1934} & 1501 \\
\hline 
\end{tabular}
\end{minipage}
\end{center}
with
\begin{center}
target:  
\begin{minipage}[c]{0.2\columnwidth}
\small
\centering
\begin{tabular}{|c|}
\hline
 count(*) \\
\hline
1\\
\hline 
\end{tabular}
\end{minipage}
$\ $ and prediction:
\begin{minipage}[c]{0.2\columnwidth}
\centering
\small
\begin{tabular}{|c|}
\hline
 count(*) \\
\hline
1\\
\hline 
\end{tabular}\,,
\end{minipage}
\end{center}
MultiTabQA identifies inter-table association of column \emph{id} of table \emph{highschooler} and column \emph{student\_id} of table \emph{likes}. It correctly disambiguates the lexical occurrence of $1934$ in columns \emph{like\_id} and \emph{student\_id} and correctly performs \emph{count}. 

A failure case also illustrates the challenges: for the question \emph{find the average weight for each pet type} with input table:
\begin{center}
\begin{minipage}[c]{1\columnwidth}
\centering
\small
\begin{tabular}{|c|c|c|c|}
\hline
PetID & PetType  & pet\_age & weight \\
\hline
 2001 & cat & 3 & 12.0\\
 2002 & dog & 2 & 13.4 \\
 2003 & dog & 1 & 9.3 \\
\hline 
\end{tabular}
\end{minipage}
\end{center}
with 
\begin{center}
target:
{\small
\begin{tabular}{|c|c|}
\hline
 avg(weight) & PetType  \\
\hline
 12.0 & cat \\
 11.35 & dog \\
\hline
\end{tabular}
}
\end{center}
and
\begin{center}
prediction:
{\small
\begin{tabular}{|c|c|}
\hline
 PetType &  avg(weight) \\
\hline
 cat & 12.0 \\
 dog & 13.4\\
  \hline
\end{tabular}\,,
}
\end{center}
MultiTabQA swaps the ordering of the 2 columns and fails to compute \emph{average} leading to an incorrect measure by table exact match. The column, row and cell metrics (precision, recall and F1) measure correctness of individual table units without measuring the ordering. Column metrics measure predicted column \emph{PetType} as correct and \emph{avg(weight)} as incorrect without measuring ordering of the 2 columns. Row \emph{cat | 12.0} is measured as correct, while \emph{dog | 13.4} is measured incorrect without measuring the ordering among them. Out of the $4$ target cells, \emph{cat}, \emph{dog}, \emph{12.0} are measured as correct.

\begin{figure}[t!]
\begin{subfigure}{0.49\columnwidth}
  \centering
  \resizebox{1\columnwidth}{!}{\includegraphics[width=1.\columnwidth]{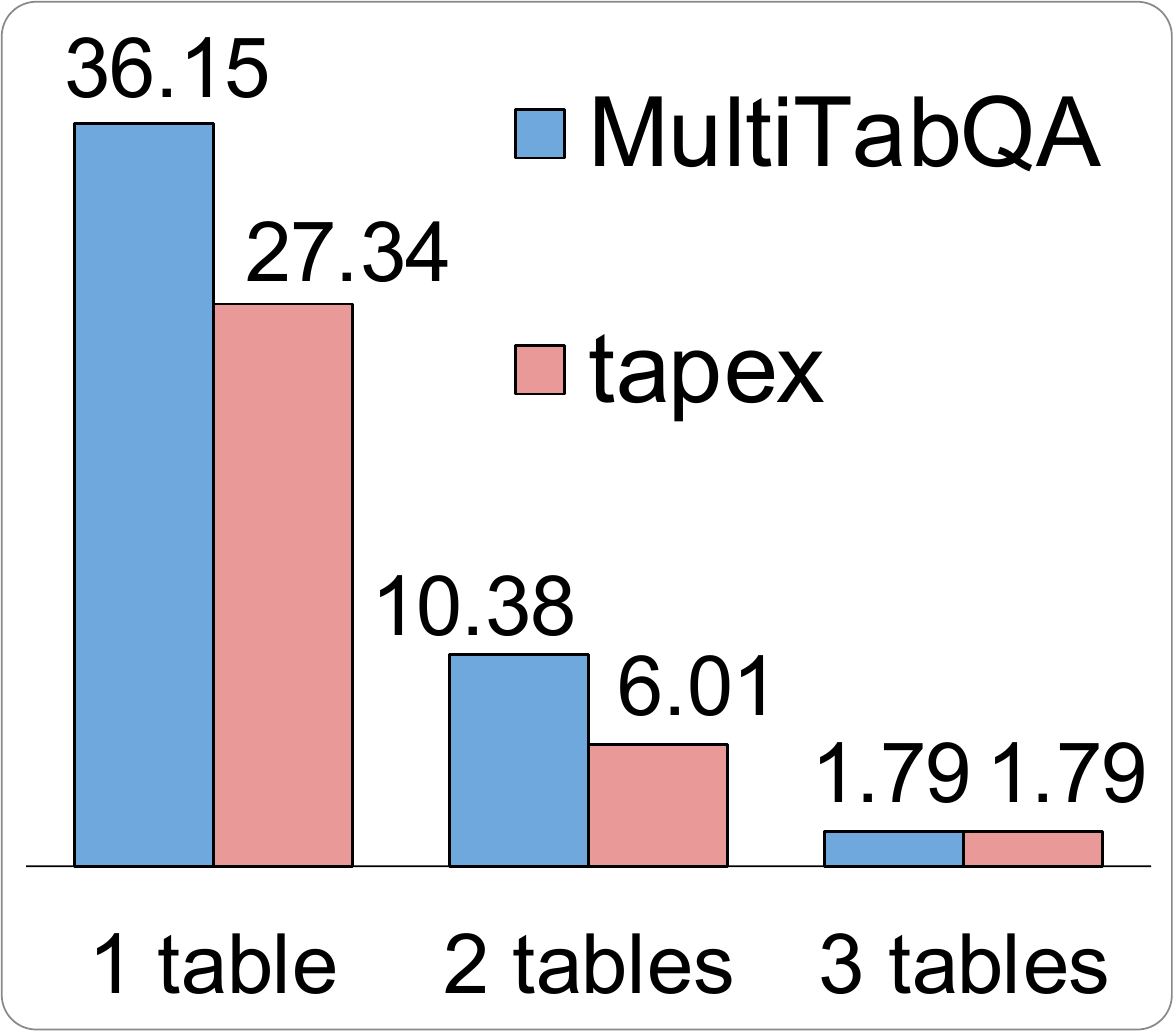}}
  \caption{Table EM Accuracy}
  \label{fig:table_exact_match}
\end{subfigure}
\begin{subfigure}{0.49\columnwidth}
  \centering
  \resizebox{1\columnwidth}{!}{\includegraphics[width=1.0\columnwidth]{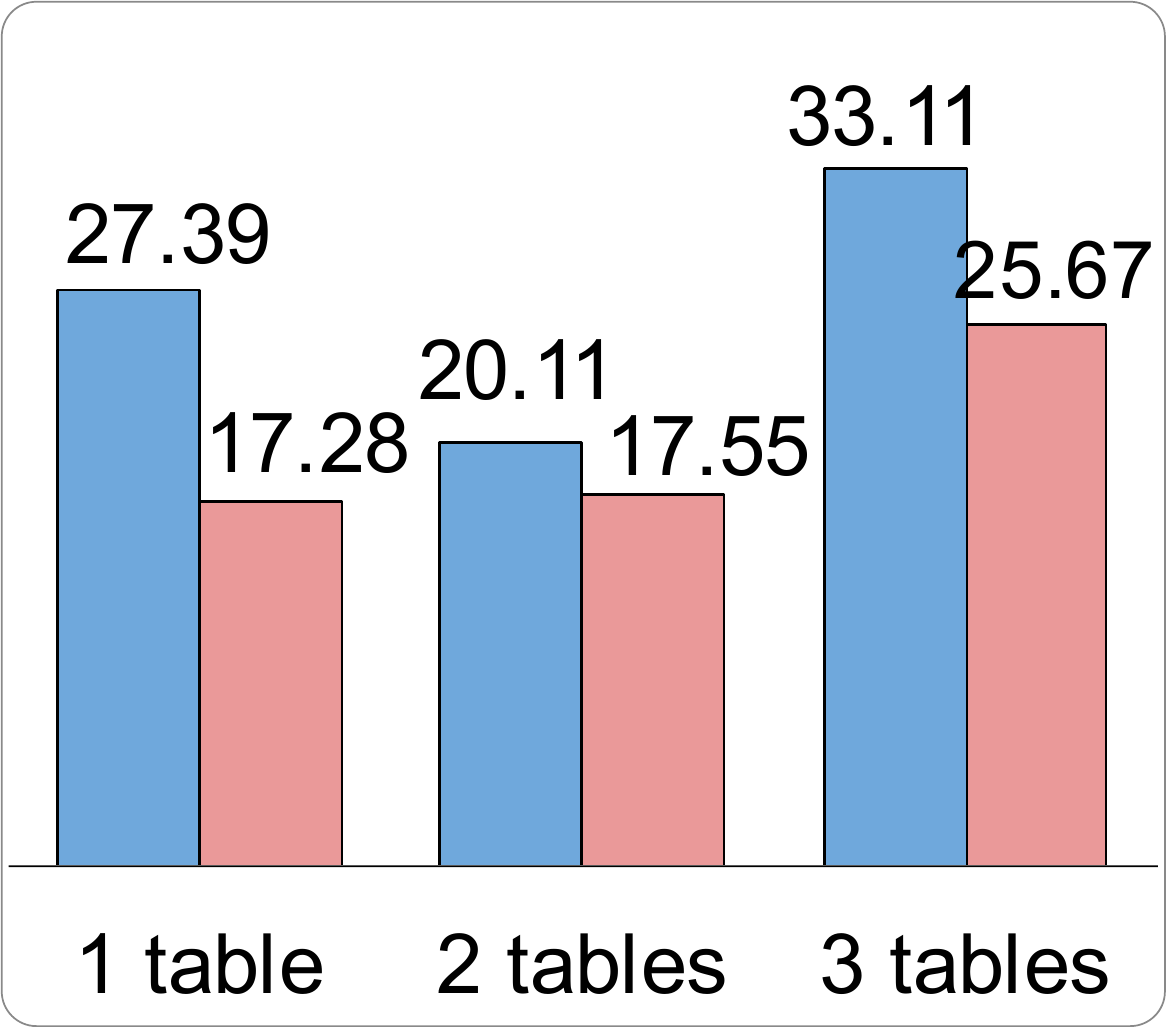}}
  \caption{Row EM F1}
  \label{fig:row_f1}
\end{subfigure}
\begin{subfigure}{0.49\columnwidth}
  \centering
  \resizebox{1\columnwidth}{!}{\includegraphics[width=1.0\columnwidth]{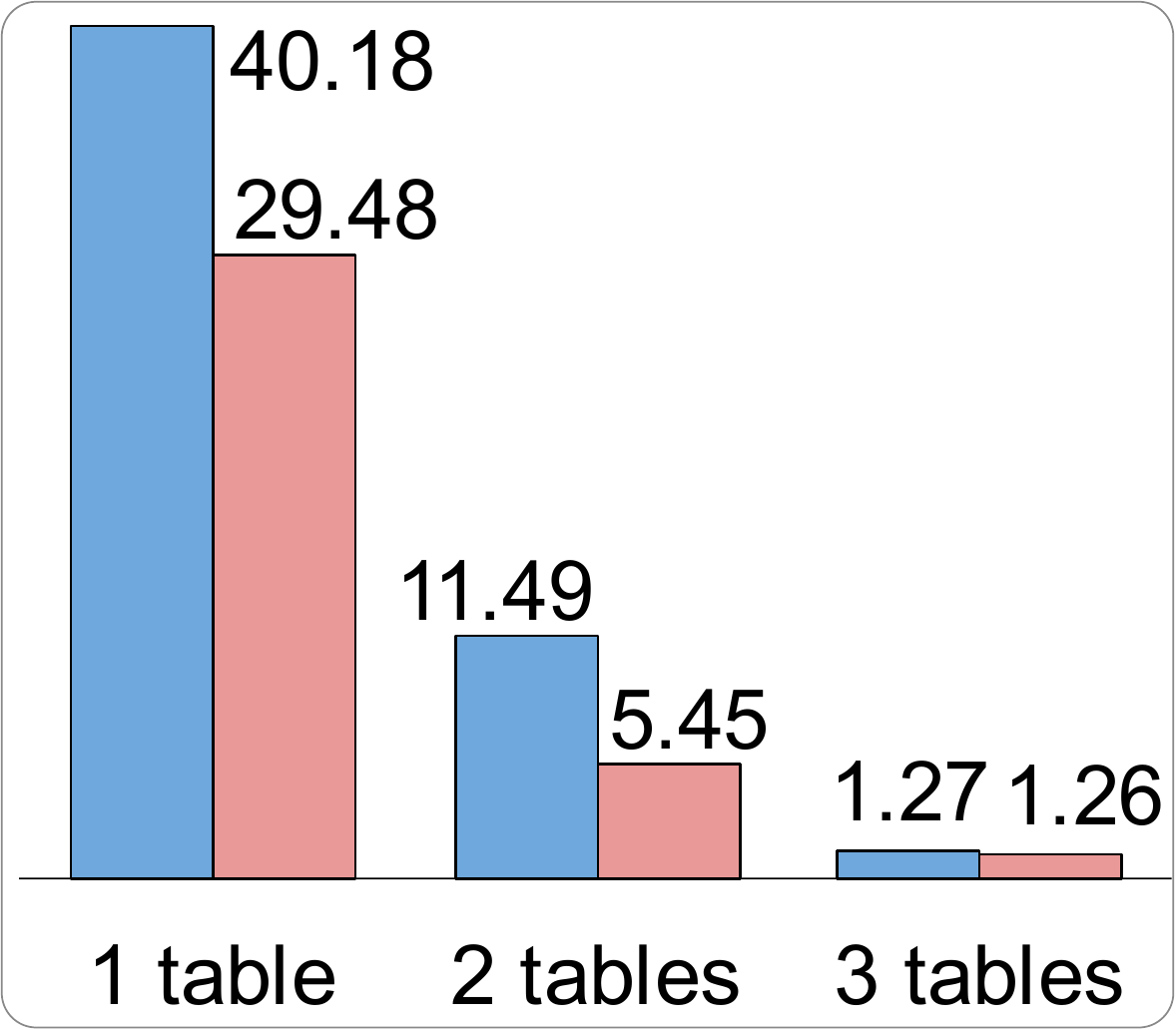}}
  \caption{Column EM F1}
  \label{fig:column_exact_match}
\end{subfigure}
\begin{subfigure}{.49\columnwidth}
  \centering
  \resizebox{1\columnwidth}{!}{\includegraphics[width=1.\columnwidth]{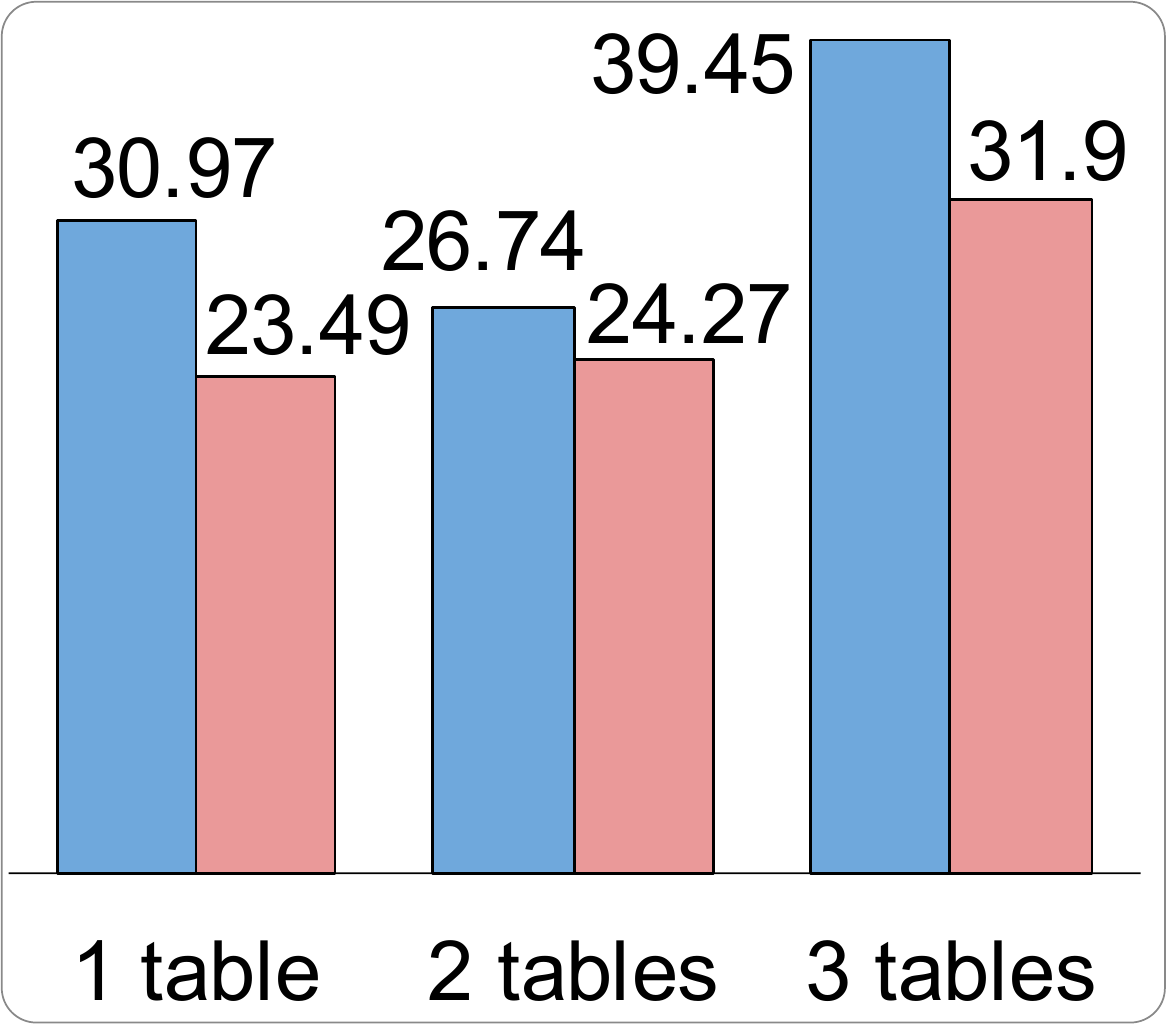}}
  \caption{Cell EM F1}
  \label{fig:cell_exact_match}
\end{subfigure}
\caption{Evaluation results on Spider evaluation samples segregated by number of input tables.}
\label{fig:segegrateinputtables}
\end{figure}

\paragraph{Impact of the number of input tables.}
The number of input tables increases the complexity of the questions and directly impacts the effectiveness of the models. We segregate evaluation on Spider validation set on the basis of number of input tables and compare the results to study the impact of input table number. We observe from Figure~\ref{fig:segegrateinputtables} that effectiveness reduces as the number of tables increases for both MultiTabQA and \texttt{tapex-base}. However, MultiTabQA fares better than \texttt{tapex-base} when the number of input tables increases. MultiTabQA generates  whole tables, rows, columns and cells better than \texttt{tapex-base} as observed in Figure~\ref{fig:table_exact_match}, \ref{fig:row_f1}, \ref{fig:column_exact_match}~and~\ref{fig:cell_exact_match}. 
The gain of MultiTabQA in table exact match for one-table context is around $8.81\%$, for two-tables context around $4.37\%$, and it performs similar to \texttt{tapex-base} for three-tables context. It also has a significant higher score on rows, columns and cells, on both single and multi-tabular context. 

We also observe that while the column and table EM decreases dramatically when using several tables (Figure~\ref{fig:table_exact_match} and \ref{fig:column_exact_match}), the row and cell EM does not (Figure~\ref{fig:row_f1} and ~\ref{fig:cell_exact_match}). This indicates that MultiTabQA can generate rows and cells as effectively in single and multiple input tables settings but fail to do so for columns and consequently for the whole table. This is due to the fact that certain  columns in the answer, particularly ones with numbers such as floats, are challenging to generate. The error from the incorrect columns propagates and are accumulated in the table EM leading to a significant drop in performance for multi-table queries.

\paragraph{Ablation on training stages.}
We perform ablation on the pre-training stages to analyse the contribution of each dataset. The simplest setting is to pre-train with Spider SQL queries, i.e., Stage~2. To evaluate the effectiveness of single table Tapex pre-training samples, the next setting comprises of stages 1 and 2, i.e., pre-train with Tapex pre-training and Spider SQL dataset. The final comparison is with the three-stage pre-training as described in Section \ref{sec:pretraining}. The results for one run of the experiments are displayed in Table~\ref{tab:exact_match_pretraining_results}. 
We observe that table exact match is highest for both pre-training and fine-tuning for the three-stage training. Stage~2 fares better than Stage~1+2 on table exact match, and generally has better precision and F1 scores but lower recall. The three-stage pre-training with our synthetic data augmented with Spider outperforms the other settings and confirms the effectiveness of our synthetic data samples in boosting model efficacy.

\section{Related work}
Tabular QA is a research direction in the broader topic of table understanding \cite{jena-etal-2022-recasting-tnli,Shigarov2022TableUP} in natural language processing. Recent advances in table representation \cite{Eisenschlos2021MATEMA} and pre-training \cite{Cheng2021FORTAPUF,Liu2022FewShotTU,Cheng2021FORTAPUF}, table fact verficiation \cite{Gu2022PASTATA,Zhou2022TablebasedFV}, table numeric reasoning \cite{Shankarampeta2022EnhancingTR,Zhou2022TaCubePD}, table-to-text generation \cite{Andrejczuk2022TableToTextGA}, text-to-table generation \cite{table2text2022}, table summarization \cite{jain-etal-2018-mixed,10.1145/2501040.2501981,Zhang:2020:SET}, and table question answering \cite{tabert2020,Zhang:2020:SET,herzig-etal-2020-tapas, zhu2021tatqa, liu2021tapex,Cheng2021HiTabAH,nan2021feta,Ma2021OpenDQ,pal-etal-2022-parameter,https://doi.org/10.48550/arxiv.2207.05270,Zhou2022TaCubePD} have shown the adaptability of language models to table processing.

\section{Conclusion}
In this work, we propose a new task of multi-table question answering without intermediate logical forms to fill the gap of existing end-to-end table QA research which focused only on single-table QA. We release a pre-training dataset of $132,645$ samples to effectively train a seq2seq model. We fine-tune and evaluate our model, MultiTabQA, on natural language questions of three datasets: Spider, GeoQuery and Atis, to test the efficacy in a multi-table setting. As many multi-table questions result in tables, we train the model to generate tables. This necessitates table-specific  metrics at various levels of granularity which we design to evaluate the effectiveness of our model. We demonstrate that such metrics is insightful in understanding model behavior. MultiTabQA outperforms existing state-of-the-art single table QA model fine-tuned to adapt to a multi-table QA setting.

\section{Limitations}
Our synthetic pre-training dataset was automatically generated from manual templates, which inspite of dataset creation scalability and low cost, may limit the diversity of the generated SQL queries. Our model, MultiTabQA, requires improvement in numeracy understanding and numeric operations. Real numbers are specially challenging and the model may not be able to correctly generate all the digits of the number correctly rending the generated cell incorrect. Furthermore, large input tables pose a challenge as the input sequence may get truncated beyond the model's maximum sequence length. This has practical limitation in the size and number of input tables which the model can accommodate before truncation which leads to incorrect answers. 

\section{Ethical Considerations}
The task and model proposed in the paper is aimed at broadening the scope of TabularQA research. All the datasets used in this research, apart from our synthetic data, are publicly available in peer-reviewed articles and referenced in this paper. The synthetic SQL dataset we release was generated over a standard benchmark database which has been annotated by 11 Yale students as mentioned in the original paper. Our synthetic samples use templates annotated by the authors of this work and do not use any user-specific data or information. We will be providing open access to our datasets for use in future research under the MIT License. All datasets, including the synthetic pre-training dataset and all datasets adapted for multi-table QA will be released. Our model is built over \texttt{tapex-base} which in turn has been trained over \texttt{bart-base}. Our work did not explicitly handle any bias which exists in the aforementioned pre-trained models.

\section{Acknowledgements}
We thank Elsevier’s Discovery Lab for their support throughout this project and funding this work. This work was also supported by the Dutch Research Council (NWO) under project numbers 016.Vidi.189.039 and 314-99-301,
by H2020-EU.3.4. Societal Challenges, 
Smart, Green and Integrated Transport (814961), 
and by the Hybrid Intelligence Center, a 10-year program funded by the Dutch Ministry of Education, Culture and Science through NWO, \url{https://hybrid-intelligence-centre.nl}. 
All content represents the opinion of the authors, which is not necessarily shared or endorsed by their respective employers and/or sponsors. 

\clearpage
\bibliographystyle{acl_natbib}
\bibliography{references}

\clearpage
\appendix


\end{document}